\documentclass[letterpaper]{article} % DO NOT CHANGE THIS
\usepackage{aaai2026}  % DO NOT CHANGE THIS
\usepackage{times}  % DO NOT CHANGE THIS
\usepackage{helvet}  % DO NOT CHANGE THIS
\usepackage{courier}  % DO NOT CHANGE THIS
\usepackage[hyphens]{url}  % DO NOT CHANGE THIS
\usepackage{graphicx} % DO NOT CHANGE THIS
\usepackage{natbib}  % DO NOT CHANGE THIS AND DO NOT ADD ANY OPTIONS TO IT
\usepackage{caption} % DO NOT CHANGE THIS AND DO NOT ADD ANY OPTIONS TO IT
\usepackage{algorithm}
\usepackage{algorithmic}
\usepackage{multirow}
\usepackage[table]{xcolor}
\usepackage{booktabs}
\usepackage{amsmath,amssymb,amsfonts}
\usepackage{makecell}
\usepackage{newfloat}
\usepackage{listings}
\DeclareCaptionStyle{ruled}{labelfont=normalfont,labelsep=colon,strut=off} % DO NOT CHANGE THIS
\floatstyle{ruled}
\newfloat{listing}{tb}{lst}{}
\floatname{listing}{Listing}
\title{CADTrack: Learning Contextual Aggregation with Deformable Alignment for Robust RGBT Tracking}
\author{
    Hao Li\textsuperscript{\rm 1,2},
    Yuhao Wang\textsuperscript{\rm 2},
    Xiantao Hu\textsuperscript{\rm 3},
    Wenning Hao\textsuperscript{\rm 1}\thanks{Corresponding authors.},
    Pingping Zhang\textsuperscript{\rm 2}\footnotemark[1],
    Dong Wang\textsuperscript{\rm 4},\\
    Huchuan Lu\textsuperscript{\rm 2,4}
}
\affiliations{
    \textsuperscript{\rm 1}College of Command and Control Engineering, Army Engineering University of PLA\\
    \textsuperscript{\rm 2}School of Future Technology, Dalian University of Technology\\
    \textsuperscript{\rm 3}School of Computer Science and Engineering, Nanjing University of Science and Technology\\
    \textsuperscript{\rm 4}School of Information and Communication Engineering, Dalian University of Technology\\
    lihao@aeu.edu.cn, 924973292@mail.dlut.edu.cn, xiantaohu@njust.edu.cn, hwnbox@aeu.edu.cn, \{zhpp,wdice,lhchuan\}@dlut.edu.cn
}

\usepackage{bibentry}
\begin{document}

\maketitle

\begin{abstract}
RGB-Thermal (RGBT) tracking aims to exploit visible and thermal infrared modalities for robust all-weather object tracking. 
However, existing RGBT trackers struggle to resolve modality discrepancies, which poses great challenges for robust feature representation. 
This limitation hinders effective cross-modal information propagation and fusion, which significantly reduces the tracking accuracy. 
To address this limitation, we propose a novel \textbf{C}ontextual \textbf{A}ggregation with \textbf{D}eformable Alignment framework called \textbf{CADTrack} for RGBT Tracking. 
To be specific, we first deploy the Mamba-based Feature Interaction (MFI) that establishes efficient feature interaction via state space models. 
This interaction module can operate with linear complexity, reducing computational cost and improving feature discrimination. 
Then, we propose the Contextual Aggregation Module (CAM) that dynamically activates backbone layers through sparse gating based on the Mixture-of-Experts (MoE). 
This module can encode complementary contextual information from cross-layer features. 
Finally, we propose the Deformable Alignment Module (DAM) to integrate deformable sampling and temporal propagation, mitigating spatial misalignment and localization drift. 
With the above components, our CADTrack achieves robust and accurate tracking in complex scenarios. 
Extensive experiments on five RGBT tracking benchmarks verify the effectiveness of our proposed method.
The source code is released at https://github.com/IdolLab/CADTrack.
\end{abstract}

\begin{figure}[t]
  \centering
  \includegraphics[width=0.90\linewidth]{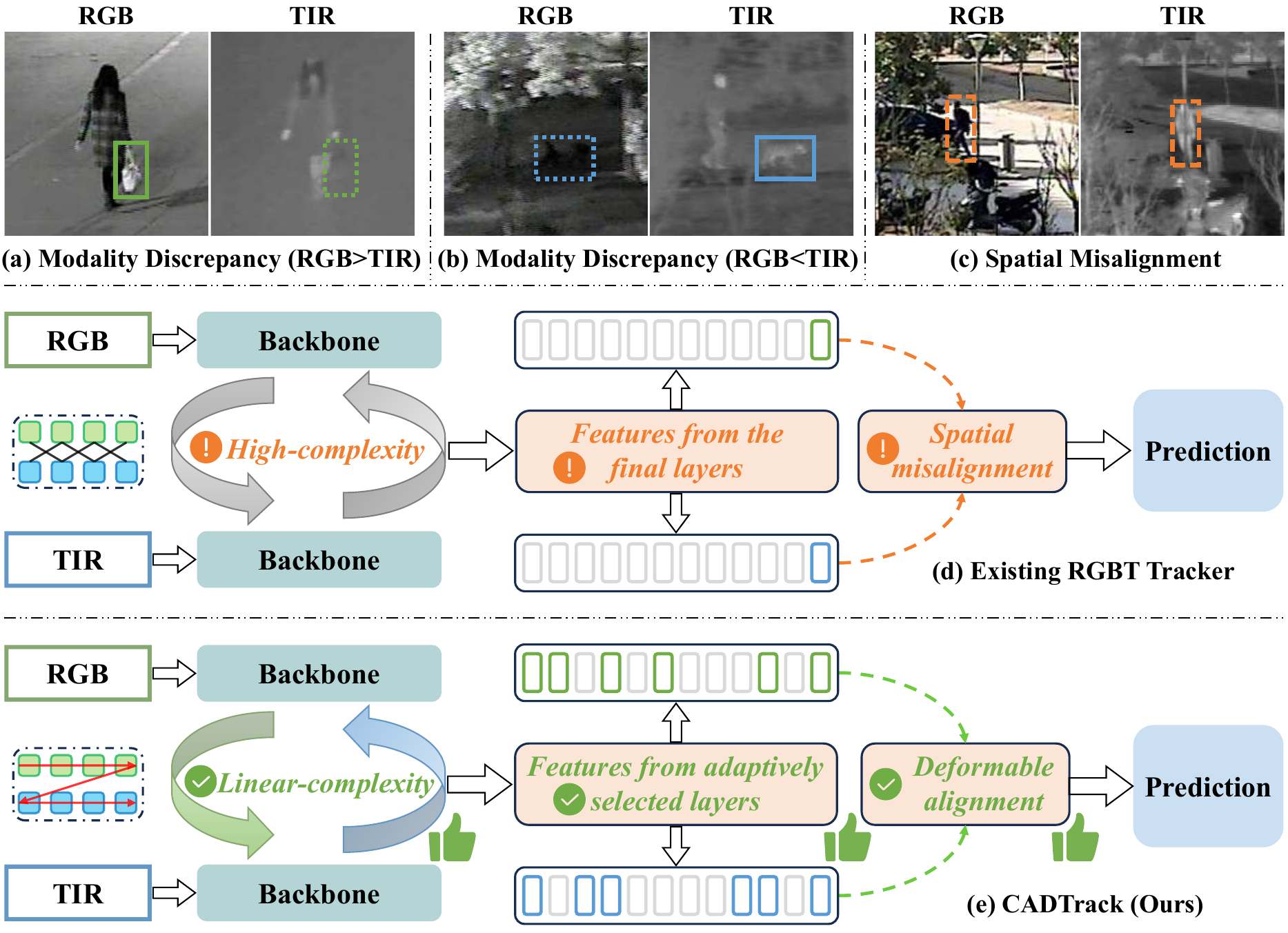}
  \caption{Comparison with different RGBT tracking paradigms. (a)-(c) The limitations of RGBT tracking include modality discrepancies and spatial misalignment. (d) Existing RGBT trackers exhibit high-complexity feature interaction, use features only from the final layers, and suffer from spatial misalignment. (e) Our framework introduces linear-complexity modality interaction, selects features from multiple layers, and employs deformable alignment.}
  \label{fig:motivation}
\end{figure}
\section{Introduction}
RGB-Thermal (RGBT) tracking is a fundamental task in computer vision and image processing.
It estimates object states by fusing complementary information from visible (RGB) and thermal infrared (TIR) modalities.
Generally, the RGB modality provides rich textures and color details under normal lighting.
The TIR modality provides structural and semantic information under lighting variations.
The fusion of these modalities can address limitations of single-modality trackers, such as failure in illumination changes, occlusions, or sensor noise.
These advantages make RGBT tracking suitable for all-day and all-weather scenarios.

In fact, most existing RGBT trackers focus on cross-modal fusion~\cite{xia2023unified,fan2024querytrack,hu2025dual,tan2025you}.
However, they overlook modality discrepancies, as shown in Fig.~\ref{fig:motivation}.
First, complex fusion methods~\cite{chen2024emoe} achieve robust information interaction but always result in high computational costs, which hinder real-time performance.
Second, prior methods~\cite{zhu2023rgbd1k,zhu2024unimod1k,zhu2024self,xu2024learning,feng2025atctrack,cai2025spmtrack,liu2025mambavlt,guo2025dreamtrack} extract features solely from the final layers of backbone networks.
They neglect cross-layer semantic information that could enhance occlusion handling and localization accuracy.
Third, modality discrepancies often lead to spatial misalignment during motion or viewpoint changes~\cite{zhang2024amnet}.
This naturally raises the question: \textbf{\textit{Can we design a RGBT tracker that dynamically adapts to modality discrepancies while ensuring computational efficiency?}}

To address the above challenges, we propose CADTrack, a novel framework that combines linear-complexity feature interaction, contextual aggregation, and deformable alignment.
%depicted
As shown in Fig.~\ref{fig:pipline}, our framework comprises three key components: Mamba-based Feature Interaction (MFI), Contextual Aggregation Module (CAM), and Deformable Alignment Module (DAM).
The MFI employs State Space Models (SSMs)~\cite{ssm1} to establish efficient feature interaction, reducing computation cost while preserving contextual coherence.
Meanwhile, the CAM dynamically activates backbone layers through sparse gating based on the Mixture-of-Experts (MoE), which achieves contextual aggregation of multi-level features.
By leveraging complementary information from shallow-level and deep-level features, it addresses the challenge of cross-layer feature under-utilization.
Finally, the DAM maintains spatial alignment through modality-specific deformable sampling and temporal propagation.
Extensive experiments on five benchmarks verify the effectiveness of our method, achieving state-of-the-art performance in both accuracy and robustness.

Our main contributions are summarized as follows:
\begin{itemize}
\item We propose CADTrack, a novel framework combining linear-complexity feature interaction, contextual aggregation, and deformable alignment for RGBT tracking.
\item We propose a Contextual Aggregation Module (CAM) that activates backbone layers through sparse gating, facilitating contextual aggregation of multi-level features.
\item We propose a Deformable Alignment Module (DAM) to solve cross-modal spatial misalignment through deformable sampling and temporal propagation.
\item Extensive experiments on five RGBT tracking benchmarks demonstrate that our method achieves state-of-the-art performance while maintaining real-time efficiency.
\end{itemize}
\section{Related Work}
\subsection{Cross-Modal Fusion in RGBT Tracking}
Recent advances have catalyzed RGBT tracking into three primary paradigms: early fusion, middle fusion and late fusion.
Early fusion means fusing two modalities in the input level.
For instance, TPF~\cite{lu2025breaking} introduces a pixel-level fusion strategy by leveraging a task-driven progressive learning framework.
Middle fusion first extracts features independently from two modalities and then fuses them at the feature level.
For example, TBSI~\cite{tbsi} employs cross-modal feature fusion through template-bridged interactions.
GMMT~\cite{gmmt} enhances feature-level fusion by leveraging generative models.
AINet~\cite{lu2025rgbt} utilizes Mamba-based feature fusion to achieve all-layer interactions.
Late fusion means aggregating tracking results from two modalities.
For example, JMMAC~\cite{9364880} implements late fusion by linearly combining modality-specific response maps.
Additionally, methods that leverage Parameter-Efficient Fine-Tuning (PEFT), such as ProTrack~\cite{yang2022prompting}, ViPT~\cite{vipt}, BAT~\cite{BAT2024}, SDSTrack~\cite{SDSTrack} and OneTracker~\cite{OneTracker}, enable RGB-to-RGBT transfer.
However, current methods suffer from high computational cost and inadequate cross-modal fusion.
Different from existing fusion methods, our method constructs a unified feature representation space that bridges heterogeneous modality characteristics with linear-complexity SSMs.
\subsection{Mixture-of-Experts in RGBT Tracking}
MoE routes inputs to specialized experts via a learnable gate, performing an adaptive weighted sum over selected experts~\cite{yuksel2012twenty}.
For RGBT tracking, MoE has been leveraged to adapt to modality-specific variations by activating experts customized for RGB or TIR features.
For example, AETrack~\cite{zhu2025adaptive} employs multiple experts for feature extraction.
MoETrack~\cite{tang2024revisiting} utilizes decision-level experts with confidence-based routing.
XTrack~\cite{tan2024xtrack} incorporates shared and modality-specialized experts for input routing.
These advances improve robustness by adapting to modality variations and optimizing the expert selection.
However, current methods ignore the adaptive fusion of multi-level features.
This limits their ability to handle occlusion and scale challenges.
Different from existing MoE-based methods, our method aggregates cross-layer features through sparse gating.
This allows a contextual aggregation of multi-level features to enhance tracking robustness.
\subsection{Temporal Modeling in RGBT Tracking}
RGBT tracking addresses dynamic targets with evolving appearances, requiring robust spatiotemporal modeling.
Previous methods rely on static templates, while recent methods aim to leverage temporal information.
For example, STMT~\cite{sun2024transformer} introduces dynamic multi-modal templates.
TATrack~\cite{TATrack} integrates temporal-aware feature fusion.
CSTrack~\cite{feng2025cstrack} unifies spatiotemporal fusion within a Vision Transformer (ViT) backbone.
STTrack~\cite{hu2025exploiting} establishes inter-frame token propagation for long-range dependencies.
MambaVT~\cite{lai2025mambavt} pioneers a Mamba-based architecture for long-range modeling.
However, current methods neglect explicit spatial misalignment.
Different from existing methods, our method resolves this limitation through spatial correction and spatiotemporal feature propagation.
\begin{figure*}
  \centering
  \includegraphics[width=.90\textwidth]{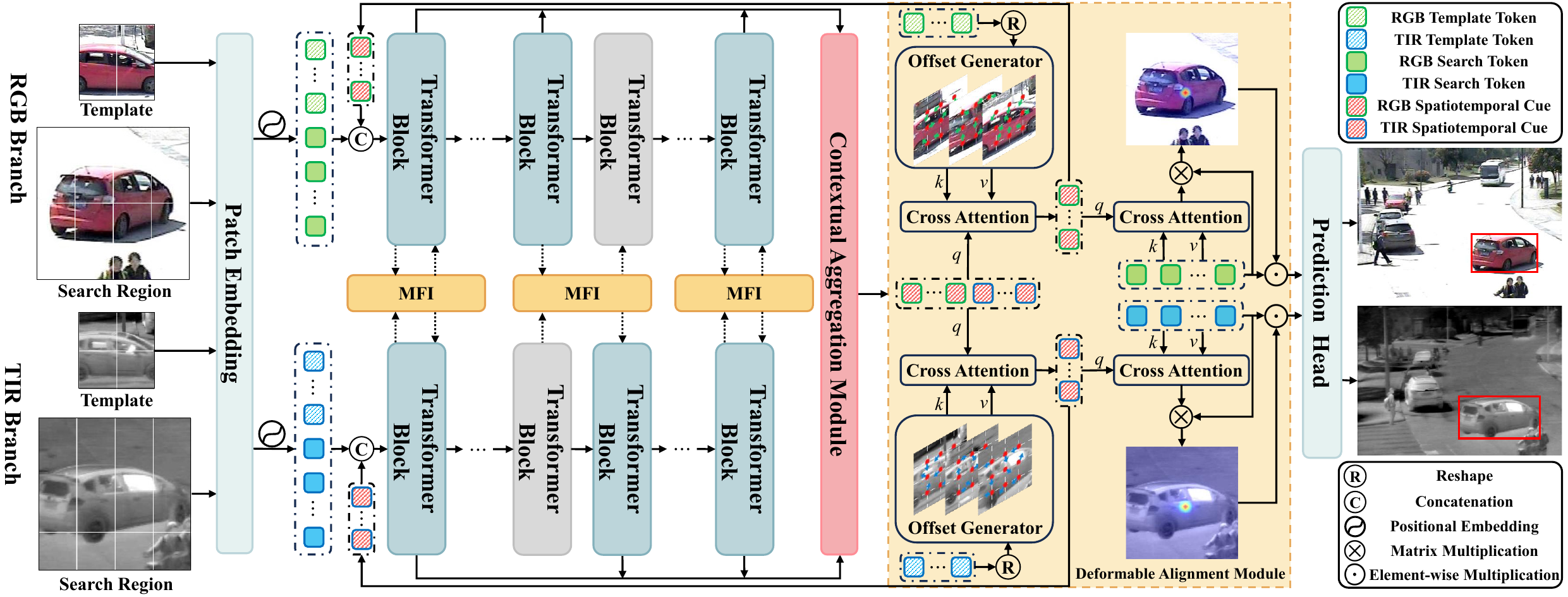}
  \caption{Overall framework of our proposed CADTrack. Firstly, input templates and search regions are tokenized with spatiotemporal alignment cues from previous frames. Then, they are fed into the backbone network with MFI for selective feature interaction. Meanwhile, CAM aggregates multi-level contextual features using modality-specific sparse gating, while DAM generates updated cues through spatial guidance for precise alignment. Finally, a prediction head is used for target localization.}
  \label{fig:pipline}
\end{figure*}
\section{Methodology}
In this paper, we propose CADTrack for RGBT tracking, which includes three key components: Mamba-based Feature Interaction (MFI), Contextual Aggregation Module (CAM) and Deformable Alignment Module (DAM).
The overall framework of CADTrack is shown in Fig.~\ref{fig:pipline}.
MFI facilitates efficient cross-modal feature interaction via SSMs, enabling selective feature fusion while maintaining linear complexity.
CAM aggregates multi-level features through sparse gating based on MoE.
DAM generates spatiotemporal alignment cues by integrating deformable sampling with temporal propagation, resolving spatial misalignment between modalities.
The overall framework and key components are described in the following subsections.
\subsection{Overall Framework}
The RGBT tracking task is initialized by a target annotation in the first image pair of RGB and TIR modalities.
At time step $t$, we process initial templates $\mathbf{Z}_m^0 \in \mathbb{R}^{3 \times H_z \times W_z}$, dynamic templates $\mathbf{Z}_m^t \in \mathbb{R}^{3 \times H_z \times W_z}$, and search regions $\mathbf{S}_m^t \in \mathbb{R}^{3 \times H_x \times W_x}$ for modalities $m \in \{R,T\}$.
Each image is split into $P \times P$ non-overlapping patches and linearly projected into $C$-dimensional tokens
$\mathbf{F}_m^{Z_0} \in \mathbb{R}^{N_z \times C}$,
$\mathbf{F}_m^{Z_t} \in \mathbb{R}^{N_z \times C}$,
$\mathbf{F}_m^{S_t} \in \mathbb{R}^{N_x \times C}$,
where $N_z$ and $N_x$ denote the numbers of patch tokens.

Then, we construct modality-specific representations by concatenating features with spatiotemporal alignment cues $\mathbf{C}_{m}^{t}$ from previous frames:
\begin{align}
\mathbf{F}_m^0 = \left[ \mathbf{C}_{m}^{t}; \mathbf{F}_m^{Z_0}; \mathbf{F}_m^{Z_t}; \mathbf{F}_m^{S_t} \right]. \label{eq:init_token}
\end{align}
These tokens are processed through Transformer blocks:
\begin{align}
\mathbf{F}_m^l = \mathcal{T}^l(\mathbf{F}_m^{l-1}),
\label{eq:transformer}
\end{align}
where $l\in \{1,2,\dots,L\}$ is the layer number. 
Each Transformer $\mathcal{T}^l$~\cite{vit} contains Multi-Head Self-Attention (MHSA) and Feed-Forward Network (FFN).

To enable efficient feature interaction, we introduce MFI between different modalities via SSMs.
For multi-level feature aggregation, we propose CAM to combine the output features from all backbone layers.
Meanwhile, we propose DAM to generate spatiotemporal alignment cues $\mathbf{C}_m^{t+1}$ for subsequent frames.
These cues enhance the aggregated features via cross-attention.
Finally, the multi-modal features $\mathbf{H}_m^t$ are concatenated and compressed via convolutional operations to reduce channels while preserving spatial resolution.
The obtained features are fed into the prediction head:
\begin{align}
{{\mathbf{B}}^t} = \phi \left({\left[ {{\mathbf{H}}_R^t;{\mathbf{H}}_T^t} \right]} \right), \label{eq:predict}
\end{align}
where $\phi$ is a fully convolutional network with stacked Conv-BN-ReLU layers, and ${\mathbf{B}}^t$ denotes the tracking result.
\begin{figure}[t]
  \centering
  \includegraphics[width=0.8\linewidth]{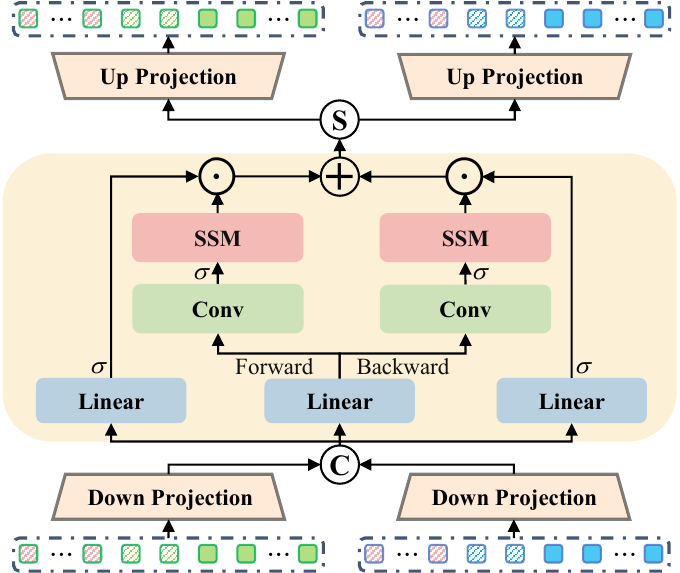}
  \caption{Details of our proposed MFI.}
  \label{fig:MambaBridge}
\end{figure}
\subsection{Mamba-based Feature Interaction}
To fully bridge RGB and TIR modalities, we propose the MFI, a cross-modal interaction module leveraging Mamba.
Traditional interaction methods suffer from high computational cost.
In contrast, our MFI adapts to modality-specific characteristics while maintaining computational efficiency.

As shown in Fig.~\ref{fig:MambaBridge}, the features $\mathbf{F}_m^l$ at the $l$-th layer are first projected into a shared latent space:
\begin{align}
\mathbf{\hat{F}}_m^l = \mathbf{W}_{m}^{down} \mathbf{F}_m^l,
\end{align}
where $\mathbf{W}_m^{down}$ is a learnable projection matrix that performs down-sampling by reducing the channel dimension.
It significantly decreases computational overhead while aligning feature representations across modalities.

Then, the projected features are concatenated into a unified sequence $\mathbf{F}_{RT}$, which is processed through the forward Mamba $\mathcal{M}_f$ and backward Mamba $\mathcal{M}_b$:
\begin{align}
\mathbf{\hat{F}}_{RT} &= \mathcal{M}_f(\mathbf{F}_{RT}) + \mathcal{M}_b(\mathbf{F}_{RT}), \\
\mathcal{M}_*(x) &= \text{SSM}_*(\sigma(\mathcal{D}(\Gamma(x)))) \odot \sigma(\Gamma(x)),
\end{align}
where $\Gamma$ is the linear projection.
$\mathcal{D}$ is the convolution operation.
$\sigma$ is the SiLU activation~\cite{elfwing2018sigmoid}.
$\text{SSM}_f$ and $\text{SSM}_b$ are the forward and backward SSM, respectively.
$\odot$ is the element-wise multiplication.

Afterwards, the features are split into $\mathbf{\tilde{F}}_m^{l}$, projected back to the original dimension via $\mathbf{W}_{m}^{up}$, and adaptively fused through residual connections:
\begin{align}
\mathbf{F}_m^{l} = \mathbf{F}_m^l + \mathbf{W}_{m}^{up} \mathbf{\tilde{F}}_m^{l}.
\end{align}

This module establishes a unified feature space for RGB and TIR modalities, enabling efficient cross-modal feature interaction.
Unlike previous methods, our MFI can enhance the robustness against feature degradation while maintaining the linear computation complexity.
\begin{figure}
  \centering
  \includegraphics[width=0.95\linewidth]{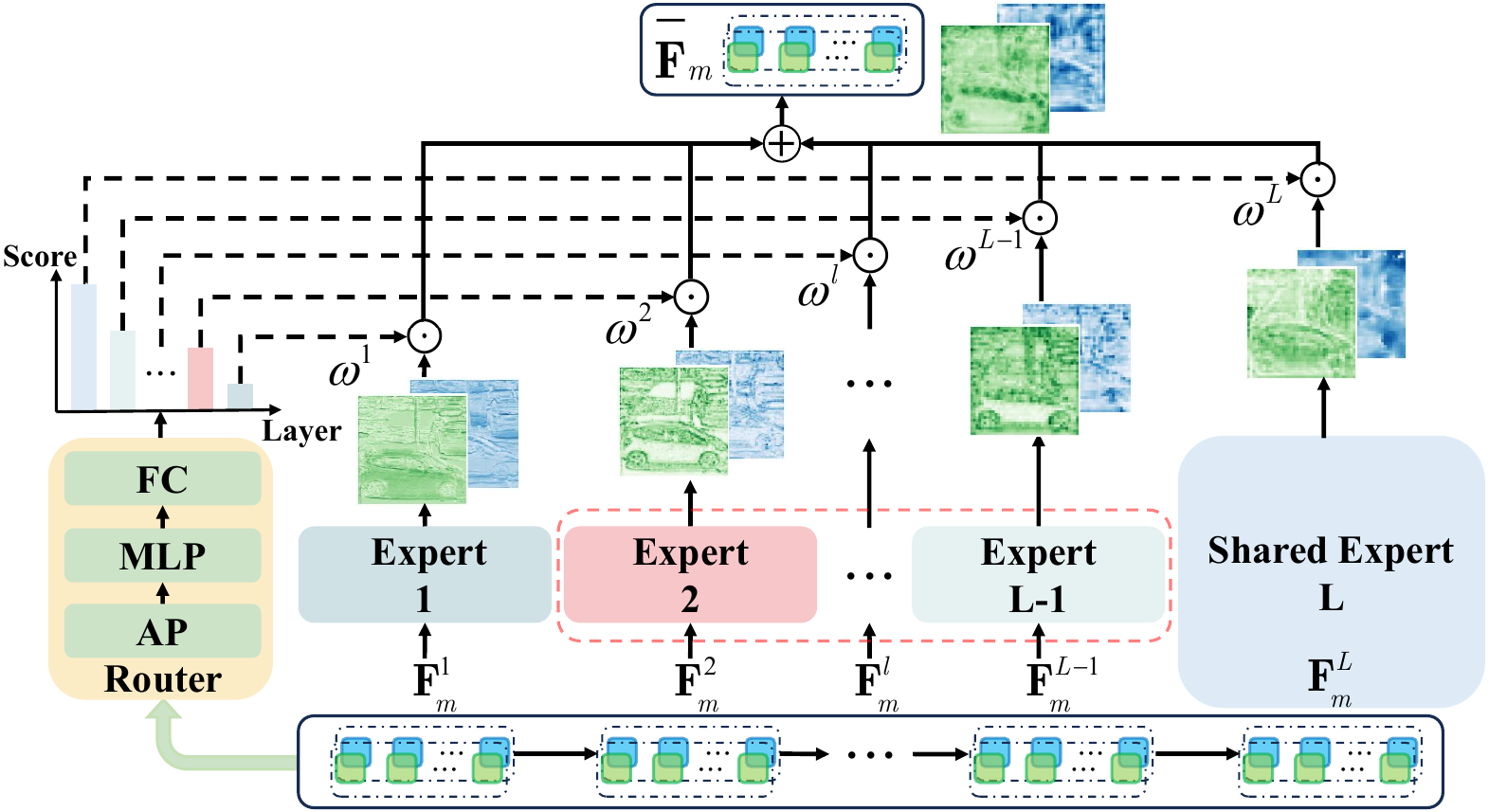}
  \caption{The structure of our proposed CAM.}
  \label{fig:CAMoE}
\end{figure}
\subsection{Contextual Aggregation Module}
Existing feature aggregation methods face two critical limitations.
First, fixed layer selection strategies~\cite{SIEVLTrack} fail to adapt to dynamic scene variations.
Second, simple aggregation strategies~\cite{lu2025rgbt} amplify noise and drift risks in RGBT tracking.
Therefore, we propose the CAM to address scene-specific challenges and achieve robust feature aggregation.
Specifically, our CAM leverages sparse activation principles to dynamically select multi-level features.

As shown in Fig.~\ref{fig:CAMoE}, our proposed CAM establishes parallel expert pools for RGB and TIR modalities.
Besides, a modality-shared router $\mathcal{R}$ generates expert selection scores ${s}_m$ by aggregating features from all $L$ Transformer layers:
\begin{equation}
{s}_m = {\rm{{\cal R}}}\left(\left[ {{\bf{F}}_m^1;{\bf{F}}_m^2; \ldots ;{\bf{F}}_m^L}\right] \right), \label{moe} \end{equation}
\begin{equation}
\mathcal{R}(x) = \mathcal{Y}(\mathcal{P}(\mathcal{A}(x))),
\end{equation}
where $\mathcal{A}$ is the Average Pooling (AP).
$\mathcal{P}$ is the Multilayer Perceptron (MLP).
$\mathcal{Y}$ is the Fully Connected (FC) layer.
The selected layers are then processed through expert networks:
\begin{equation}
\tilde{\mathbf{F}}_m^l = \mathbf{W}^l \mathbf{F}_m^l,
\end{equation}
where $\mathbf{W}^l$ denotes modality-shared projection weights.
The linear projection mitigates feature distribution gaps across different layers~\cite{liu2023improving}.

Our CAM is implemented based on three key insights: (1) The shallowest expert ($l=1$) is always activated to preserve high-frequency spatial details essential for object deformation; (2) The deepest expert ($l=L$) is selected as the shared expert to maintain semantic information during contextual aggregation; (3) The remaining experts are selected via top-$k$ sparse activation based on router scores $s_m$.

The resulting modality-specific aggregation is performed over the selected expert set $\mathcal{E}_m$ using learnable weights $w^l$ for selected layers, as follows:
\begin{equation}
\overline{\mathbf{F}}_m = \sum_{l \in \mathcal{E}_m} w^l \odot \tilde{\mathbf{F}}_m^l.
\end{equation}

Unlike static feature aggregation methods, our proposed CAM enables context-aware expert selection that adapts to scene variations while maintaining computational efficiency.
\begin{figure}[t]
  \centering
  \includegraphics[width=0.96\linewidth]{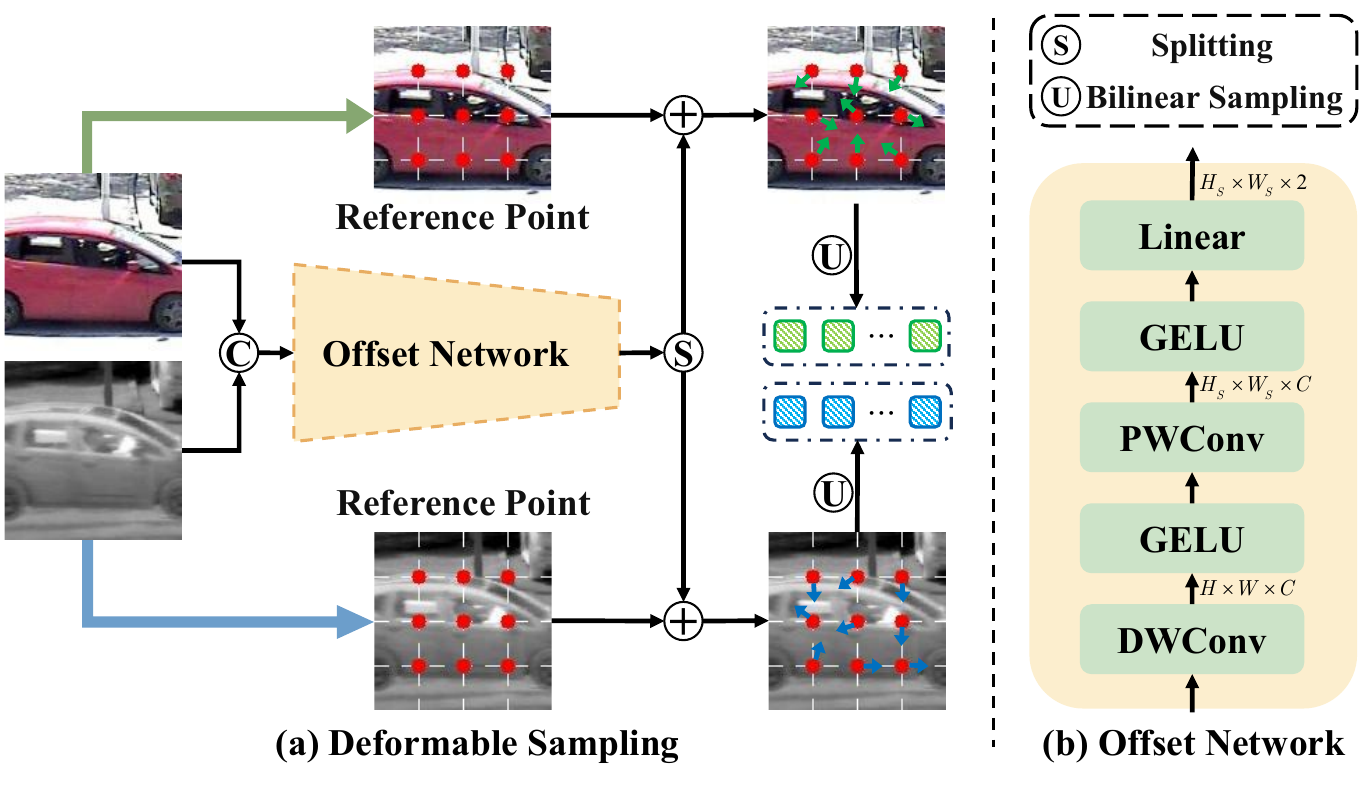}
  \caption{Deformable alignment of DAM.}
  \label{fig:offset}
\end{figure}
\subsection{Deformable Alignment Module}
To address the spatial misalignment caused by modality discrepancies, we propose the DAM, which generates spatiotemporal alignment cues $\mathbf{C}_m^t \in \mathbb{R}^{N_K \times C}$ via deformation sampling and temporal propagation.

As shown in Fig.~\ref{fig:offset}, we first extract patch tokens from initial template features and dynamic template features, and reshape them into spatial representations $ {\bf{\hat F}}_m^{Z_0} \in \mathbb{R}^{H \times W \times C}$, $ {\bf{\hat F}}_m^{Z_t} \in \mathbb{R}^{H \times W \times C}$, where $H = H_z/P$ and $W = W_z/P$.
Then, these features are concatenated and processed through a convolutional mixer $\mathcal{O}$:
\begin{align}
{\bf\hat{F}}_{A} = \mathcal{O}\left( \left[ {\bf{\hat F}}_m^{Z_0}; {\bf{\hat F}}_m^{Z_t} \right] \right),\\
\mathcal{O}(x) = \delta(\mathcal{W}(\delta(\mathcal{Q}(x))),
\end{align}
where $\mathcal{W}$ and $\mathcal{Q}$ are the Point-Wise Convolution (PWConv) and Depth-Wise Convolution (DWConv), respectively.
$\delta$ is the GELU activation~\cite{hendrycks2016gaussian}.
\begin{table*}[t]
    \centering
     \caption{Performance on four RGBT tracking benchmarks. The best results are in bold.}
    \setlength{\tabcolsep}{1.3mm}{
    \renewcommand\arraystretch{0.85}{
	\resizebox{\linewidth}{!}{
	\begin{tabular}{c|c|c|c|cc|cc|cc|ccc}
        \toprule
		\multirow{2}{*}{Method} & \multirow{2}{*}{Source} &\multirow{2}{*}{Backbone}&\multirow{2}{*}{Pretraining} & \multicolumn{2}{c|}{GTOT} & \multicolumn{2}{c|}{RGBT210} & \multicolumn{2}{c|}{RGBT234} & \multicolumn{3}{c}{LasHeR} \\\cline{5-13}
		 & & & & \rule{0pt}{1.8ex}MPR$\uparrow$ & MSR$\uparrow$ & PR$\uparrow$ & SR$\uparrow$ & MPR$\uparrow$ & MSR$\uparrow$ & PR$\uparrow$  & NPR$\uparrow$  & SR$\uparrow$  \\
		\midrule

        DAPNet~\cite{zhu2019dense} & ACM MM 2019 & VGG$-$M & ImageNet & 88.2 & 70.7 & $-$ & $-$ & 76.6& 53.7 & 43.1  & 38.3  & 31.4  \\
        MANet~\cite{long2019multi} & ICCVW 2019 & VGG$-$M & ImageNet & 89.4 & 72.4 & $-$ & $-$ & 77.7& 53.9 & 45.5  & $-$  & 32.6  \\
        CMPP~\cite{wang2020cross} & CVPR 2020 & VGG$-$M & ImageNet & 92.6 & 73.8 & $-$ & $-$ & 82.3& 57.5 & $-$ & $-$  & $-$  \\
        CAT~\cite{li2020challenge} & ECCV 2020 & VGG$-$M & ImageNet & 88.9 & 71.7 & 79.2  & 53.3 & 80.4& 56.1 & 45.0  & 39.5  & 31.4  \\
        APFNet~\cite{xiao2022attribute} & AAAI 2022 & VGG$-$M & ImageNet & 90.5 & 73.7 & $-$  & $-$ & 82.7& 57.9 & 50.0  & 43.9  & 36.2  \\
        HMFT~\cite{Zhang_CVPR22_VTUAV} & CVPR 2022 & ResNet$-$50 & ImageNet & 91.2 & 74.9 & 78.6 & 53.5 & 78.8& 56.8 & $-$  & $-$  & $-$  \\
        ProTrack~\cite{yang2022prompting} & ACM MM 2022 & ViT$-$B & SOT & $-$ & $-$ & $-$ & $-$ & 78.6& 58.7 & 50.9  & $-$  & 42.1  \\
        QAT~\cite{liu2023quality} & ACM MM 2023 & ResNet$-$50 & ImageNet & 91.5 & 75.5 & 86.8 & 61.9 & 88.4& 64.3 & 64.2  &59.6 & 50.1  \\
        CMD~\cite{zhang2023efficient} & CVPR 2023 &  ResNet$-$50 & ImageNet & 89.2 & 73.4 & $-$ & $-$ & 82.4& 58.4 & 59.0  &54.6 & 46.4  \\
        ViPT~\cite{vipt} & CVPR 2023 &  ViT$-$B & SOT& $-$ & $-$ & $-$ & $-$ & 83.5 & 61.7 & 65.1 & $-$ & 52.5 \\
        TBSI~\cite{tbsi} & CVPR 2023 &  ViT$-$B & SOT& $-$ & $-$ & 85.3 & 62.5 & 87.1 & 63.7 & 69.2 & 65.7 & 55.6  \\
        STMT~\cite{sun2024transformer} & TCSVT 2024  & ViT$-$B & SOT& $-$ & $-$ & 83.0 & 59.5 & 86.5 & 63.8 & 67.4 & 63.4 & 53.7  \\
        TATrack~\cite{TATrack} & AAAI 2024  & ViT$-$B & SOT & $-$ & $-$ & 85.3 & 61.8 & 87.2 & 64.4 & 70.2 & 66.7 & 56.1  \\
        BAT~\cite{BAT2024} & AAAI 2024  & ViT$-$B & SOT & $-$ & $-$ & $-$ & $-$ & 86.8 & 64.1 & 70.2& $-$ & 56.3  \\
        GMMT~\cite{gmmt} & AAAI 2024  & ViT$-$B & SOT & $-$ & $-$ & $-$ & $-$ & 87.9 & 64.7 & 70.7 & 67.0 & 56.6  \\
        {Un-Track}~\cite{Un-Track} & CVPR 2024  & ViT$-$B & SOT & $-$ & $-$ & $-$ & $-$ & 83.7 & 61.8 & 66.7& $-$ & 53.6  \\
        SDSTrack~\cite{SDSTrack} & CVPR 2024  & ViT$-$B & SOT & $-$ & $-$ & $-$ & $-$ & 84.8 & 62.5 & 66.5& $-$ & 53.1 \\
        OneTracker~\cite{OneTracker} & CVPR 2024  & ViT$-$B & ImageNet & $-$ & $-$ & $-$ & $-$ & 85.7 & {64.2} & 67.2& $-$ & 53.8  \\
        CKD~\cite{ckd} & ACM MM 2024 & ViT$-$B & DropMAE & 93.2 & 77.2 & 88.4 & 65.2 & 90.0 & 67.4 & 73.2& 69.3 & 58.1  \\
    TPF~\cite{lu2025breaking}  &  ARXIV 2025 & ViT$-$B & DropMAE & 94.3 & 76.3 & 88.0 & 63.8 & 89.7 & 67.1 & 75.1& 71.3 & 59.5  \\
    AINet~\cite{lu2025rgbt}  &  AAAI 2025 & ViT$-$B & DropMAE & $-$ & $-$ & 86.8 & 64.1 & 89.1 & 66.8 & 73.0 &69.0 & 58.2  \\
    STTrack~\cite{hu2025exploiting} &  AAAI 2025 & ViT$-$B & SOT & $-$ & $-$ &$-$ & $-$ & 89.8 & 66.7 & 76.0 & $-$ & 60.3 \\
    CAFormer~\cite{xiao2025cross}  &  AAAI 2025 & ViT$-$B & SOT & 91.8 & 76.9 & 85.6 & 63.2 & 88.3 & 66.4 & 70.0 &66.1& 55.6 \\
    MambaVT~\cite{lai2025mambavt}  &  TCSVT 2025  & Vim$-$S & ImageNet & 94.1 & 75.3 & 88.0 & 63.7 & 88.9 & 65.8 & 73.0 & 69.5 & 57.9  \\
    XTrack~\cite{tan2024xtrack}  &  ICCV 2025 & ViT$-$B  &  DropMAE & $-$ & $-$ & $-$ & $-$ & 87.4 & 64.9 & 69.1 & $-$ & 55.7 \\
    \hline
    \rowcolor[gray]{0.92}
    \rule{0pt}{1.8ex}CADTrack &  Ours  & ViT$-$B & SOT & 95.3 & 77.8 & 88.7 & 62.9 & 90.9 & 65.6 & 75.8 & 71.9 & 60.2 \\
    \rowcolor[gray]{0.92}
    CADTrack &  Ours & ViT$-$B & DropMAE & \textbf{95.8} & \textbf{78.3} & \textbf{91.2} & \textbf{65.4} & \textbf{92.8} & \textbf{67.7} & \textbf{77.7} & \textbf{73.3} & \textbf{61.3} \\
\bottomrule
       \end{tabular}}}}
\label{overall_result}
\end{table*}

With reference points $\tilde P \in \mathbb{R}^{H_S \times W_S \times 2}$ from convolution centers, we predict modality-specific offsets to accommodate the spatial misalignment:
\begin{align}
\Delta \tilde P_m^{\tau} &= v \cdot \mathcal{G}_m^{\tau}({\bf\hat{F}}_{A}), 
\end{align}
where $\tau \in \{Z_0, Z_t\}$. $\mathcal{G}_m^{\tau}$ is a linear projection layer and $v$ is the offset magnitude.
Then, we utilize bilinear sampling $\mathcal{U}$ to amplify discriminative features~\cite{xia2022vision}:
\begin{equation}
{\bf{F}}_m^{\tau} = \mathcal{U}\left( {\bf\hat{F}}_m^{\tau}, \tilde P + \Delta \tilde P_m^{\tau} \right).
\end{equation}

Afterwards, we concatenate the sampled features to obtain ${\bf{F}}_S = \left[ {\bf{F}}_m^{Z_0}; {\bf{F}}_m^{Z_t} \right]$.
For temporal propagation, we update alignment cues via cross-modal cross-attention $\Phi$:
\begin{equation}
\mathbf{C}_m^{t+1} = \mathbf{C}_m^{t} + \Phi\left( \mathbf{C}_m^{t}, {\bf{F}}_S \right).
\end{equation}

Finally, we introduce the intra-modal cross-attention to refine features of search regions by three sequential steps: spatial guidance, feature enhancement and response generation.
The corresponding formulations are as follows:
\begin{align}
\mathbf{\hat{C}}_m^{t+1} &= \mathbf{C}_m^{t+1} + \Phi(\mathbf{C}_m^{t+1}, \overline{\mathbf{F}}_m^{S_t}), \\
\mathbf{\tilde{C}}_m^{t+1} &= \mathbf{\hat{C}}_m^{t+1} + \text{FFN}(\mathbf{\hat{C}}_m^{t+1}), \\
\mathbf{H}_m^t &= \overline{\mathbf{F}}_m^{S_t} \otimes (\mathbf{\tilde{C}}_m^{t+1})^T \odot \overline{\mathbf{F}}_m^{S_t},
\end{align}
where $\otimes$ denotes the matrix multiplication.

Unlike previous methods that rely on spatial alignment, our DAM generates spatiotemporal alignment cues through deformable sampling and temporal propagation.
It preserves modality-specific spatial characteristics while enhancing complementary information.
In addition, by leveraging modality-specific offsets and cross-modal cross-attention, our DAM addresses spatial misalignment caused by modality discrepancies without explicit supervision.
\section{Experiments}
\subsection{Implementation Details}
We implement CADTrack based on the PyTorch toolbox and train it on 4 NVIDIA V100 GPUs, with a global batch size of 32.
The framework adopts a ViT-B backbone initialized from SOT~\cite{ostrack} and DropTrack~\cite{wuyang2023dropmae}.
Training employs the AdamW optimizer~\cite{AdamW} with a learning rate $10^{-4}$ and a weight decay $10^{-4}$.
Our MFI is deployed at the 4-th, 7-th, and 10-th layers of ViT-B, employing channel compression with 8.
We select 6 experts per modality.
$N_K=1$ and $v=5$ to balance deformation adaptability and localization precision.
Template images and search regions utilize fixed resolutions of 128$\times$128 and 256$\times$256, respectively.
For training, we use the training set of LasHeR~\cite{li2021lasher}.
On VTUAV~\cite{Zhang_CVPR22_VTUAV}, we instead train exclusively on its training set.
\subsection{Comparison with State-of-the-Art Trackers}
We evaluate CADTrack on GTOT~\cite{li2016gtot}, RGBT210~\cite{Li17rgbt210}, RGBT234~\cite{li2019rgb234}, LasHeR and VTUAV benchmarks using Precision Rate (PR) and Success Rate (SR), with additional Normalized Precision Rate (NPR) for LasHeR.
Notably, ground truth annotations are misaligned in GTOT, RGBT234 and VTUAV.
Following previous evaluation practices~\cite{hu2025adaptive,shao2025pura} , we employ Maximum Precision Rate (MPR) and Maximum Success Rate (MSR) on these benchmarks.

\textbf{GTOT.}
As shown in Tab.~\ref{overall_result}, our method shows outstanding performance with 95.8\% MPR and 78.3\% MSR.
It outperforms CAFormer~\cite{xiao2025cross} by a margin of +4.0\% MPR and shows improved robustness than the early-fusion method TPF~\cite{lu2025breaking} with a margin of +2.0\% MSR.

\textbf{RGBT210.}
As shown in Tab.~\ref{overall_result}, our method achieves 91.2\% PR and 65.4\% SR.
Specifically, it outperforms TATrack~\cite{TATrack} by 5.9\% in PR through cross-modal feature interaction, while surpassing STMT~\cite{sun2024transformer} by 5.9\% in SR, validating the effectiveness of our method in handling modality-specific challenges.

\textbf{RGBT234.}
As shown in Tab.~\ref{overall_result}, our method achieves leading performance, surpassing recent trackers: +5.4\% MPR over XTrack~\cite{tan2024xtrack} and +5.2\% MSR over SDSTrack~\cite{SDSTrack}.
These results demonstrate that our method has an outstanding cross-modal feature representation capability, particularly in handling challenging scenarios with modality discrepancies.

\begin{figure}[t]
  \centering
  \includegraphics[width=0.9\linewidth]{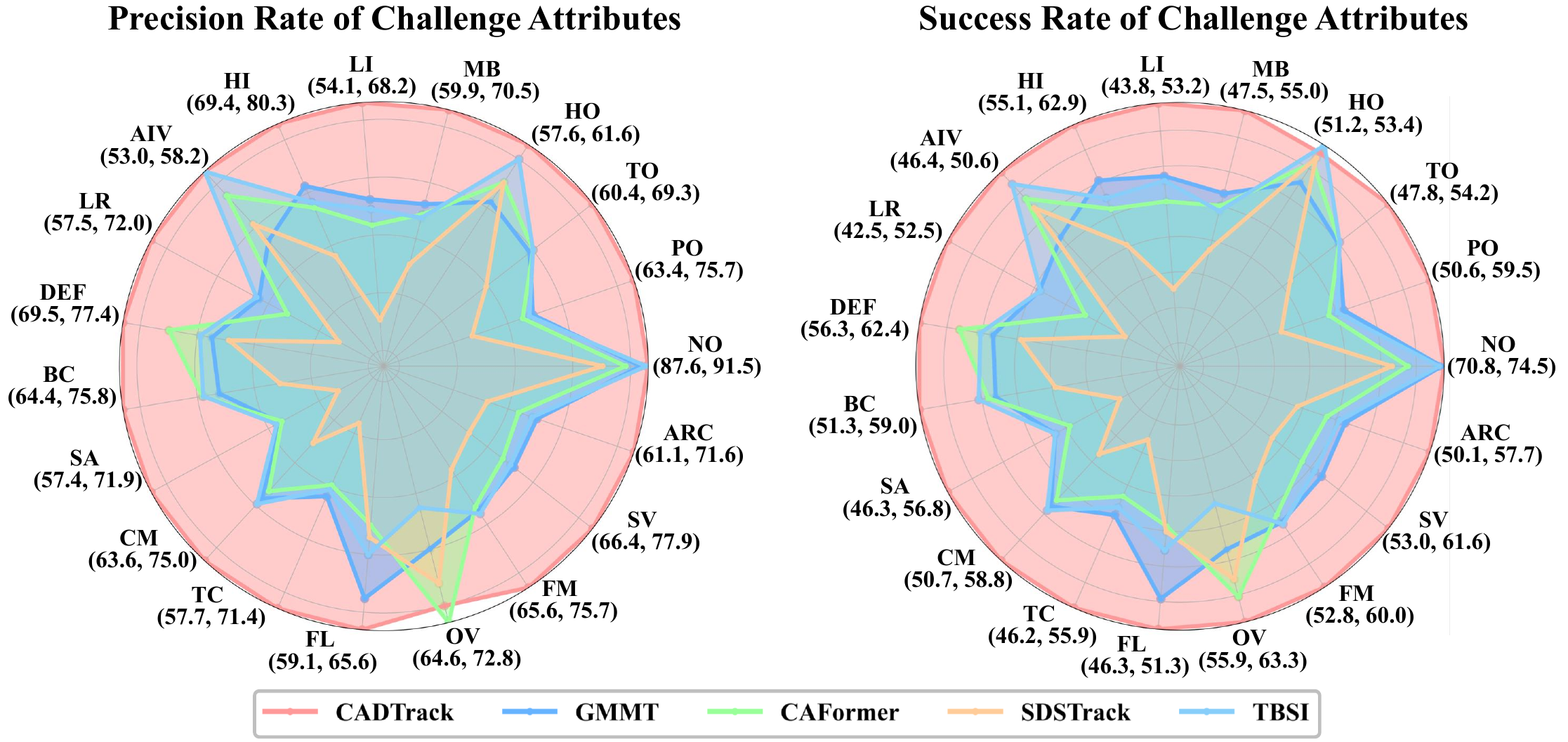}
  \caption{Attribute-based evaluation on the LasHeR dataset.}
  \label{fig:radar_plot}
\end{figure}

\textbf{LasHeR.}
As shown in Tab.~\ref{overall_result}, our method delivers state-of-the-art performance with PR 77.7\% and SR 61.3\%.
These results demonstrate notable improvements: +7.0\% PR over GMMT~\cite{gmmt}, and +7.7\% SR over Un-Track~\cite{Un-Track}.
Fig.~\ref{fig:radar_plot} further reveals the attribute-specific dominance: +10.6\% PR in motion blur (MB) and +6.1\% SR in deformation (DEF).
Notably, +13.7\% PR in thermal crossover (TC) and +9.4\% SR in low illumination (LI) highlight the robustness under challenging conditions.

\textbf{VTUAV.}
As shown in Tab.~\ref{tab:vtuav_simplified}, on the short-term (ST) subset, our method achieves 78.2\% MSR, outperforming AINet~\cite{lu2025rgbt} by +3.7\%.
On the long-term (LT) subset, it attains 61.3\% MPR, surpassing HMFT\_LT~\cite{Zhang_CVPR22_VTUAV} by +7.7\%.
These results highlight a balanced accuracy-efficiency across both ST and LT scenarios.
%--------------------------------------------
\begin{table}[!t]
\centering
\caption{Performance on VTUAV benchmark.}
\label{tab:vtuav_simplified}
\setlength{\tabcolsep}{4pt}
\renewcommand\arraystretch{0.9}{
\resizebox{0.9\linewidth}{!}{
\begin{tabular}{c|cc|cc}
\toprule
\multirow{2}{*}{Method} & \multicolumn{2}{c|}{VTUAV-ST} & \multicolumn{2}{c}{VTUAV-LT} \\
\cline{2-5}
&\rule{0pt}{1.8ex} MPR $\uparrow$ & MSR $\uparrow$ & MPR $\uparrow$ & MSR  $\uparrow$ \\
\midrule
DAFNet~\cite{gao2019deep} & 62.0 & 45.8 & 25.3 & 18.8 \\
mfDiMP~\cite{zhang2019multi} & 67.3 & 55.4 & 31.5 & 27.2 \\
HMFT~\cite{Zhang_CVPR22_VTUAV} & 75.8 & 62.7 & 41.4 & 35.5 \\
HMFT\_LT~\cite{Zhang_CVPR22_VTUAV} & - & - & 53.6 & 46.1 \\
QAT~\cite{liu2023quality} & 80.1 & 66.7 & - & - \\
CKD~\cite{ckd} & 90.2 & 77.8 & - & - \\
AINet~\cite{lu2025rgbt} & 87.1 & 74.5 & - & - \\
CAFormer~\cite{xiao2025cross} & 88.6 & 76.2 & - & - \\
MambaVT~\cite{lai2025mambavt} & 88.6 & 75.7 & - & - \\
\hline
\rowcolor[gray]{0.92}
\rule{0pt}{1.8ex}CADTrack(Ours) & \textbf{90.4} & \textbf{78.2} & \textbf{61.3} & \textbf{53.7} \\
\bottomrule
\end{tabular}
}
}
\end{table}
%------------------------------------------------------
\begin{table}[!t]
\centering
\caption{Ablation study of key components on LasHeR.}
\label{tab:ablation}
\renewcommand\arraystretch{0.9}{
\resizebox{0.84\linewidth}{!}{
\begin{tabular}{ccccc}
\toprule
Method & Pretraining & PR$\uparrow$ & NPR$\uparrow$ & SR$\uparrow$ \\
\midrule
Baseline & SOT & 67.9 & 64.4 & 54.5 \\
+ Template update & SOT & 69.1 & 65.6 & 55.6 \\
+ DAM & SOT & 72.7 & 69.3 & 58.3 \\
+ MFI & SOT &74.1 & 70.4 & 59.2 \\
+ CAM & SOT  & 75.8 & 71.9 & 60.2 \\
\rowcolor[gray]{0.92}
Full model & DropMAE  & \textbf{77.7} & \textbf{73.3} & \textbf{61.3} \\
\bottomrule
\end{tabular}
}
}
\end{table}
%---------------------------------------------
\subsection{Ablation Studies}
In this subsection, we conduct ablation experiments to assess the effect of different components.
\begin{table}[t]
\centering
\caption{Comparison of interaction mechanisms on LasHeR.}
\label{tab:fusion_type}
\renewcommand\arraystretch{0.9}{
\resizebox{0.9\linewidth}{!}{
\begin{tabular}{cccccccc}
\toprule
Module & PR$\uparrow$ & NPR$\uparrow$ & SR$\uparrow$ & Params$ \downarrow $ & FLOPs$ \downarrow $& FPS$ \uparrow $ \\
\midrule
TBSI & 74.5 & 70.9 & 59.7 & 229.2M & 108.8G & 29\\
BAT  & 75.9 & 72.1 & 60.4 & 130.1M & 77.4G & 39\\
BSI  & 76.9 & 72.6 & 60.8 & 133.1M & 79.8G & 35\\
\rowcolor[gray]{0.92}
MFI & \textbf{77.7} & \textbf{73.3} & \textbf{61.3} & \textbf{130.0M} & \textbf{77.3G} & \textbf{40}\\
\bottomrule
\end{tabular}
}
}
\end{table}
%----------------------------------------------------------
\begin{table}[!t]
\centering
\caption{Impact of interaction positions on LasHeR.}
\label{tab:position}
\renewcommand\arraystretch{0.9}{
\resizebox{0.6\linewidth}{!}{
\begin{tabular}{cccccc}
\toprule
 4 & 7 & 10 & PR$\uparrow$ & NPR$\uparrow$ & SR$\uparrow$ \\
\midrule
  \checkmark & & & 75.2 & 71.4 & 59.8  \\
  \checkmark & \checkmark & &  75.9 & 71.7 & 60.3 \\
  \rowcolor[gray]{0.92}
  \checkmark & \checkmark & \checkmark & \textbf{77.7} & \textbf{73.3} & \textbf{61.3} \\
\bottomrule
\end{tabular}
}
}
\end{table}
%---------------------------------------------------

\textbf{Component Analysis.}
As shown in Tab.~\ref{tab:ablation}, the baseline (ViT-B+convolutional fusion) achieves 67.9\% PR, revealing fundamental limitations.
Template update mechanisms address temporal drift by refreshing target representations, yielding a +1.2\% PR. 
Our DAM resolves spatial misalignment through deformable sampling, delivering the gain of +3.6\% PR.
Our MFI enables an efficient modality interaction, resulting in +1.4\% PR.
Finally, our CAM dynamically activates backbone layers through sparse gating, achieving 75.8\% PR.
The use of pre-trained DropTrack provides additional gains, showing the generalization of our method.

\textbf{Interaction Mechanism Analysis.}
We further validate the efficacy of MFI against other paradigms:
(1) template-bridged search region interaction (TBSI)~\cite{tbsi};
(2) bi-directional adapter (BAT)~\cite{BAT2024};
(3) background suppression interactive (BSI)~\cite{hu2025exploiting}.
As shown in Tab.~\ref{tab:fusion_type}, our MFI achieves an efficiency-performance balance, outperforming TBSI by +3.2\% PR with 29\% lower FLOPs and 38\% higher FPS.
This advantage stems from our linear-complexity feature interaction.
\begin{table}[t]
\centering
\caption{Effect of Mamba layers on LasHeR.}
\label{tab:depth}
\renewcommand\arraystretch{0.95}{
\begin{tabular}{ccccc}
\toprule
Layer & PR$\uparrow$ & NPR$\uparrow$ & SR$\uparrow$  \\
\midrule
1 & 75.4 & 71.4 & 60.1 \\
\rowcolor[gray]{0.92}
2 & \textbf{77.7} & \textbf{73.3} & \textbf{61.3}  \\
3  & 77.1 & 72.6 & 61.0 \\
\bottomrule
\end{tabular}
}
\end{table}
%------------------------------------------------------------
\begin{table}[t]
\centering
\caption{Expert selection strategies of CAM on LasHeR.}
\label{tab:came_num}
\renewcommand\arraystretch{0.95}{
\begin{tabular}{cccc}
\toprule
Selection Strategy & PR$\uparrow$ & NPR$\uparrow$ & SR$\uparrow$ \\
\midrule
Top-$k$ ($k=4$) & 75.5 & 71.4 & 59.9 \\
\rowcolor[gray]{0.92}
Top-$k$ ($k=6$) & \textbf{77.7} & \textbf{73.3} & \textbf{61.3} \\
Top-$k$ ($k=8$) & 74.3 & 70.2 & 58.9 \\
Fixed interval & 75.8 & 71.6 & 59.9 \\
Manual selection & 75.3 & 71.2 & 59.6 \\
\bottomrule
\end{tabular}
}
\end{table}
%------------------------------------------------------------
\begin{table}[!t]
\centering
\caption{Impact of cue quantity on LasHeR.}
\label{tab:cada_cue}
\renewcommand\arraystretch{0.9}{
\begin{tabular}{ccccc}
\toprule
$N_K$ & PR$\uparrow$ & NPR$\uparrow$ & SR$\uparrow$ & $\Delta$FLOPs\\
\midrule
\rowcolor[gray]{0.92}
1 & \textbf{77.7} & \textbf{73.3} & \textbf{61.3} & $-$\\
2 & 75.6 & 71.3 & 59.9 & +0.3\% \\
4 & 76.9 & 72.6 & 60.8 & +0.8\% \\
\bottomrule
\end{tabular}
}
\end{table}
%-----------------------------------------------------------
\textbf{Effect of Interaction Positions.}
As shown in Tab.~\ref{tab:position}, the interactions of Layers 4, 7, and 10 achieves the best performance with 77.7\% PR, demonstrating that early interactions capture spatial patterns while deeper interactions incorporate semantic abstractions.
Our cross-layer design yields complementary advantages.

\textbf{Effect of Mamba Layers in MFI.}
As shown in Tab.~\ref{tab:depth}, the 2-layer configuration achieves optimal results with 77.7\% PR.
In contrast, the 1-layer variant shows a significant performance drop in PR by 2.3\%.
The 3-layer variant exhibits a minor performance decline in PR by 0.6\%.
This slight degradation suggests that the excessive model complexity may lead to over-fitting.
\begin{figure}[tp]
  \centering
  \includegraphics[width=0.94\linewidth]{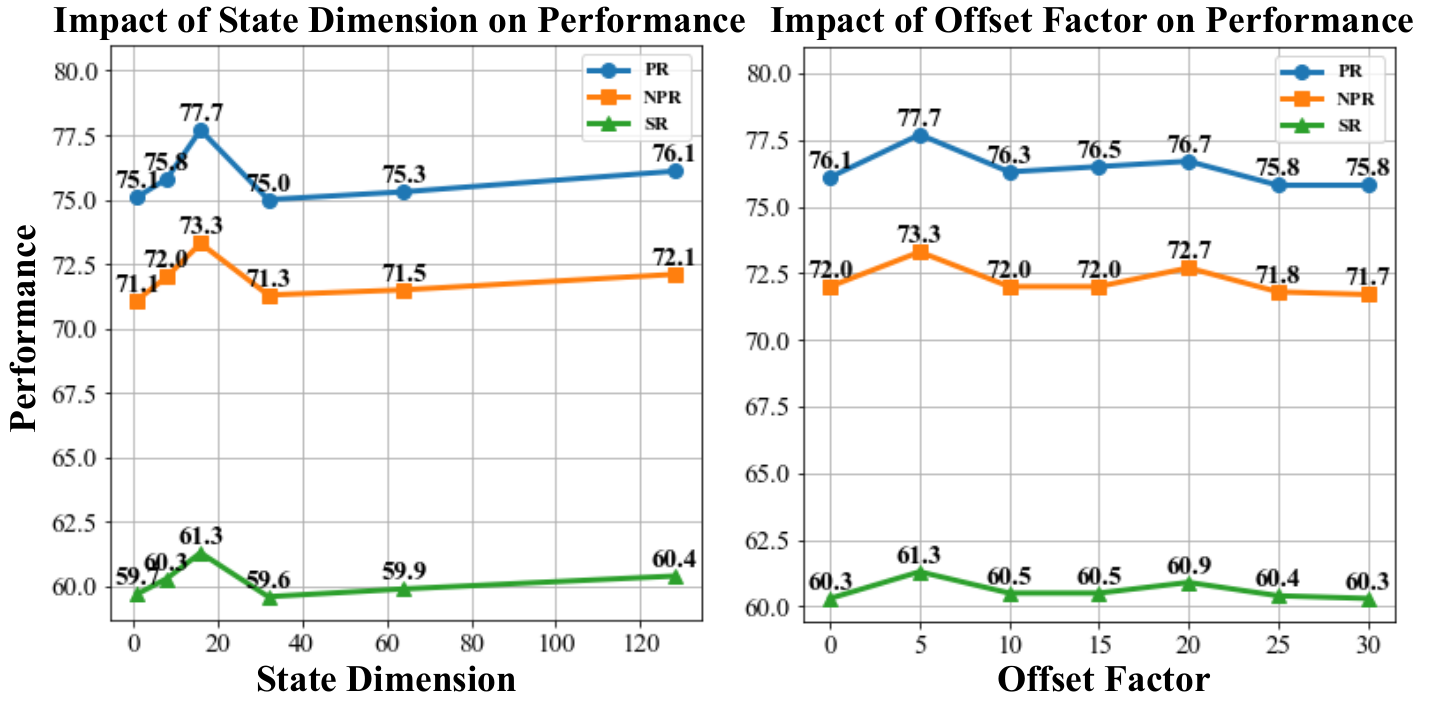}
  \caption{Comparison with different hyper-parameters.}
  \label{fig:impact}
\end{figure}
%------------------------------------------------------
\begin{figure}[tp]
  \centering
  \includegraphics[width=0.91\linewidth]{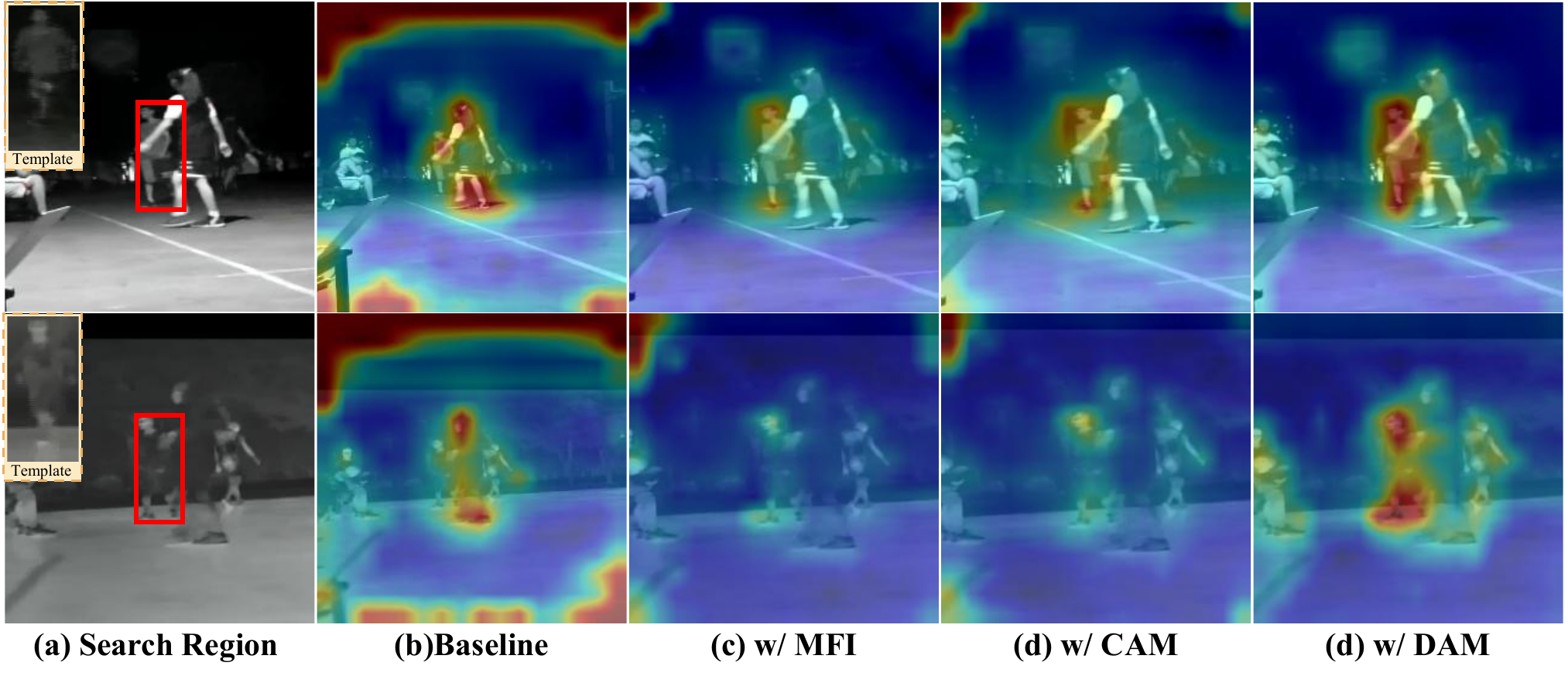}
  \caption{Attention evolution for RGB (top) and T (bottom).}
  \label{fig:attention}
\end{figure}

\textbf{Effect of State Dimensions in MFI.}
The state dimension in SSM plays a key role in modeling long-range dependencies.
As shown in Fig.~\ref{fig:impact}, the 16-dimensional configuration achieves the best performance with 77.7\% PR.
Smaller dimensions exhibit a reduced representational ability due to insufficient context encoding.
Larger dimensions perform worse despite more parameters.
These findings indicate that large state dimensions introduce redundancy, while insufficient dimensions limit context modeling.
The 16-dimensional state balances expressiveness and computational efficiency, capturing essential spatiotemporal dependencies for cross-modal feature interaction.

\textbf{Effect of Expert Numbers in CAM.}
As shown in Tab.~\ref{tab:came_num}, our method achieves optimal performance of 77.7\% PR by selecting 6 experts.
It achieves higher performance than the one with $k=4$, while avoiding the increased noise observed at $k=8$ which yields 74.3\% PR.
Compared methods show clear limitations: fixed-interval selection (every 2 layers) yields 75.8\% PR due to the lack of adaptability in dynamic scenes, while the manual selection (1,3,6,9,12) obtains 75.3\% PR due to the suboptimal spatial-semantic balance.
Our method enables dynamic prioritization: shallow layers are activated for deformation scenarios, while deep layers focus on occlusion robustness.
By leveraging this layer-specific activation, the model addresses diverse tracking challenges with improved efficiency and accuracy.

\textbf{Effect of Cue Quantity in DAM.}
As evidenced in Tab.~\ref{tab:cada_cue}, the quantity of spatiotemporal alignment cues influences tracking precision.
The optimal performance is achieved at $N_K=1$ with 77.7\% PR, significantly outperforming multi-cue configurations.
The performance difference arises from two key factors: (1) excessive cues introduce conflicting signals due to over-parameterization, and (2) single-cue processing relies on temporal consistency to simplify state representations.
Additionally, the configuration of $N_K=1$ reduces the computational overhead of multi-cue variants.

\textbf{Effect of the Offset Factor in DAM.}
The offset factor $v$ determines the deformation range.
As shown in Fig.~\ref{fig:impact}, $v=5$ achieves the best performance with 77.7\% PR.
Smaller values reduce the representation ability due to limited deformation ranges.
In contrast, larger values introduce instability during rapid motion.
These results highlight that large deformation ranges create inconsistent movement, while insufficient ranges limit adaptability.
The configuration of $v=5$ maintains tracking stability under motion variations.
\subsection{Visualization Analysis}
Fig.~\ref{fig:attention} illustrates the evolution of attention maps.
The baseline model exhibits inconsistent attention, leading to wrong target localization.
Our MFI aligns cross-modal features, enabling correct target localization.
Our CAM focuses on key features, while DAM eliminates noise for robust tracking.
\section{Conclusion}
In this work, we propose a novel feature learning framework named CADTrack for RGBT tracking.
For modality interaction, we introduce Mamba-based Feature Interaction (MFI), which achieves linear-complexity feature interaction while preserving contextual coherence.
To enhance feature robustness, we propose the Contextual Aggregation Module (CAM), dynamically activating backbone layers through sparse gating and leveraging complementary information from cross-layer features.
Furthermore, we propose the Deformable Alignment Module (DAM), resolving modality-specific misalignment via deformable sampling and temporal propagation.
Extensive experiments on five benchmarks validate the effectiveness of our method.
\section{Acknowledgments}
This work was supported in part by the National Natural Science Foundation of China (No.62576069), Dalian Science and Technology Innovation Fund (No.2023JJ11CG001) and Natural Science Foundation of Liaoning Province (No.2025-MS-025).
\bibliography{aaai2026}

@article{yuksel2012twenty,
  title={Twenty years of mixture of experts},
  author={Yuksel, Seniha Esen and Wilson, Joseph N and Gader, Paul D},
  journal={IEEE TNNLS},
  volume={23},
  number={8},
  pages={1177--1193},
  year={2012}
}

@inproceedings{AdamW,
  author    = {Ilya Loshchilov and Frank Hutter},
  title     = {Decoupled Weight Decay Regularization},
  booktitle = {ICLR},
  year      = {2019}
}

@inproceedings{BAT2024,
  author    = {Bing Cao and Junliang Guo and Pengfei Zhu and Qinghua Hu},
  title     = {Bi-directional Adapter for Multimodal Tracking},
  booktitle = {AAAI},
  volume    = {38},
  pages     = {927--935},
  year      = {2024}
}

@inproceedings{ckd,
  author    = {Andong Lu and Jiacong Zhao and Chenglong Li and Yun Xiao and Bin Luo},
  title     = {Breaking Modality Gap in {RGBT} Tracking: Coupled Knowledge Distillation},
  booktitle = {ACM MM},
  pages     = {9291--9300},
  year      = {2024}
}

@inproceedings{feng2025cstrack,
  author    = {Xiaokun Feng and Dailing Zhang and Shiyu Hu and Xuchen Li and Meiqi Wu and Jing Zhang and Xiaotang Chen and Kaiqi Huang},
  title     = {{CSTrack}: Enhancing {RGB-X} Tracking via Compact Spatiotemporal Features},
  booktitle = {ICML},
  year      = {2025}
}

@inproceedings{gmmt,
  author    = {Zhangyong Tang and Tianyang Xu and Xiaojun Wu and Xue-Feng Zhu and Josef Kittler},
  title     = {Generative-based Fusion Mechanism for Multi-modal Tracking},
  booktitle = {AAAI},
  volume    = {38},
  pages     = {5189--5197},
  year      = {2024}
}

@inproceedings{hu2025exploiting,
  author    = {Xiantao Hu and Ying Tai and Xu Zhao and Chen Zhao and Zhenyu Zhang and Jun Li and Bineng Zhong and Jian Yang},
  title     = {Exploiting Multimodal Spatial-Temporal Patterns for Video Object Tracking},
  booktitle = {AAAI},
  volume    = {39},
  pages     = {3581--3589},
  year      = {2025}
}

@article{hu2025dual,
  author    = {Yufan Hu and Zekai Shao and Bin Fan and Hongmin Liu},
  title     = {Dual-Level Modality De-Biasing for {RGBT} Tracking},
  journal   = {IEEE TIP},
  volume    = {34},
  number    = {},
  pages     = {2667-2679},
  year      = {2025}
}

@article{lai2025mambavt,
  author    = {Simiao Lai and Chang Liu and Jiawen Zhu and Ben Kang and Yang Liu and Dong Wang and Huchuan Lu},
  title     = {{MambaVT}: Spatio-temporal Contextual Modeling for Robust {RGBT} Tracking},
  journal   = {IEEE TCSVT},
  volume    = {35},
  number    = {9},
  pages     = {9312-9323},
  year      = {2025}
}

@article{li2016gtot,
  author    = {Chenglong Li and Hui Cheng and Shiyi Hu and Xiaobai Liu and Jin Tang and Liang Lin},
  title     = {Learning Collaborative Sparse Representation for Grayscale-Thermal Tracking},
  journal   = {IEEE TIP},
  volume    = {25},
  number    = {12},
  pages     = {5743--5756},
  year      = {2016}
}

@inproceedings{Li17rgbt210,
  author    = {Chenglong Li and Nan Zhao and Yijuan Lu and Chengli Zhu and Jin Tang},
  title     = {Weighted Sparse Representation Regularized Graph Learning for {RGBT} Object Tracking},
  booktitle = {ACM MM},
  pages     = {1856--1864},
  year      = {2017}
}

@article{li2019rgb234,
  author    = {Chenglong Li and Xinyan Liang and Yijuan Lu and Nan Zhao and Jin Tang},
  title     = {RGBT Object Tracking: Benchmark and Baseline},
  journal   = {PR},
  volume    = {96},
  pages     = {106977},
  year      = {2019}
}

@article{li2021lasher,
  author    = {Chenglong Li and Wanlin Xue and Yaqing Jia and Zhichen Qu and Bin Luo and Jin Tang and Dengdi Sun},
  title     = {{LasHeR}: A Large-scale High-diversity Benchmark for {RGBT} Tracking},
  journal   = {IEEE TIP},
  volume    = {31},
  pages     = {392--404},
  year      = {2021}
}

@inproceedings{liu2023improving,
  author    = {Yuan Liu and Songyang Zhang and Jiacheng Chen and Zhaohui Yu and Kai Chen and Dahua Lin},
  title     = {Improving Pixel-based {MIM} by Reducing Wasted Modeling Capability},
  booktitle = {ICCV},
  pages     = {5361--5372},
  year      = {2023}
}

@article{lu2025breaking,
  author    = {Andong Lu and Yuanzhi Guo and Wanyu Wang and Chenglong Li and Jin Tang and Bin Luo},
  title     = {Breaking Shallow Limits: Task-Driven Pixel Fusion for Gap-free {RGBT} Tracking},
  journal   = {arXiv preprint arXiv:2503.11247},
  year      = {2025}
}

@inproceedings{lu2025rgbt,
  author    = {Andong Lu and Wanyu Wang and Chenglong Li and Jin Tang and Bin Luo},
  title     = {{RGBT} Tracking via All-layer Multimodal Interactions with Progressive Fusion {Mamba}},
  booktitle = {AAAI},
  volume    = {39},
  pages     = {5793--5801},
  year      = {2025}
}

@inproceedings{OneTracker,
  author    = {Lingyi Hong and Shilin Yan and Renrui Zhang and Wanyun Li and Xinyu Zhou and Pinxue Guo and Kaixun Jiang and Yiting Chen and Jinglun Li and Zhaoyu Chen and others},
  title     = {{OneTracker}: Unifying Visual Object Tracking with Foundation Models and Efficient Tuning},
  booktitle = {CVPR},
  pages     = {19079--19091},
  year      = {2024}
}

@inproceedings{ostrack,
  author    = {Botao Ye and Hong Chang and Bingpeng Ma and Shiguang Shan and Xilin Chen},
  title     = {Joint Feature Learning and Relation Modeling for Tracking: A One-stream Framework},
  booktitle = {ECCV},
  pages     = {341--357},
  year      = {2022}
}

@article{SIEVLTrack,
  author    = {Ning Li and Bineng Zhong and Qihua Liang and Zhiyi Mo and Jian Nong and Shuxiang Song},
  title     = {{SIEVL-Track}: Exploring Semantic Information Enhancement for Visual-Language Object Tracking},
  journal   = {IEEE TCSVT},
  volume    = {35},
  number    = {6},
  pages     = {5872--5884},
  year      = {2025}
}

@inproceedings{SDSTrack,
  author    = {Xiaojun Hou and Jiazheng Xing and Yijie Qian and Yaowei Guo and Shuo Xin and Junhao Chen and Kai Tang and Mengmeng Wang and Zhengkai Jiang and Liang Liu and others},
  title     = {{SDSTrack}: Self-Distillation Symmetric Adapter Learning for Multi-modal Visual Object Tracking},
  booktitle = {CVPR},
  pages     = {26551--26561},
  year      = {2024}
}

@inproceedings{shao2025pura,
  author    = {Zekai Shao and Yufan Hu and Bin Fan and Hongmin Liu},
  title     = {{PURA}: Parameter Update-Recovery Test-Time Adaption for {RGBT} Tracking},
  booktitle = {CVPR},
  pages     = {22089--22098},
  year      = {2025}
}

@inproceedings{ssm1,
  author    = {Albert Gu and Isys Johnson and Karan Goel and Khaled Saab and Tri Dao and Atri Rudra and Christopher R{\'e}},
  title     = {Combining Recurrent, Convolutional, and Continuous-Time Models with Linear State Space Layers},
  booktitle   = {NeurIPS},
  volume    = {34},
  pages     = {572--585},
  year      = {2021}
}

@article{sun2024transformer,
  author    = {Dengdi Sun and Yajie Pan and Andong Lu and Chenglong Li and Bin Luo},
  title     = {Transformer {RGBT} Tracking with Spatio-temporal Multimodal Tokens},
  journal   = {IEEE TCSVT},
  volume    = {34},
  number    = {11},
  pages     = {12059-12072},
  year      = {2024}
}

@inproceedings{TATrack,
  author    = {Hongyu Wang and Xiaotao Liu and Yifan Li and Meng Sun and Dian Yuan and Jing Liu},
  title     = {Temporal Adaptive RGBT Tracking with Modality Prompt},
  booktitle = {AAAI},
  volume    = {38},
  pages     = {5436--5444},
  year      = {2024}
}

@inproceedings{tan2024xtrack,
  author    = {Yuedong Tan and Zongwei Wu and Yuqian Fu and Zhuyun Zhou and Guolei Sun and Eduard Zamfi and Chao Ma and Danda Pani Paudel and Luc Van Gool and Radu Timofte},
  title     = {{XTrack}: Multimodal Training Boosts {RGB-X} Video Object Trackers},
  booktitle = {ICCV},
  pages={5734--5744},
  year      = {2025}
}

@article{tang2024revisiting,
  author    = {Tang, Zhangyong and Xu, Tianyang and Wu, Xiao-Jun and Zhu, Xuefeng and Cheng, Chunyang and Feng, Zhenhua and Kittler, Josef},
  title     = {Revisiting rgbt tracking benchmarks from the perspective of modality validity: A new benchmark, problem, and solution},
  journal   = {IEEE TIP},
  volume    = {34},
  number    = {},
  pages     = {7235-7249},
  year      = {2025}
}

@inproceedings{tbsi,
  author    = {Tianrui Hui and Zizheng Xun and Fengguang Peng and Junshi Huang and Xiaoming Wei and Xiaolin Wei and Jiao Dai and Jizhong Han and Si Liu},
  title     = {Bridging Search Region Interaction With Template for {RGBT} Tracking},
  booktitle = {CVPR},
  pages     = {13630--13639},
  year      = {2023}
}

@inproceedings{Un-Track,
  author    = {Zongwei Wu and Jilai Zheng and Xiangxuan Ren and Florin-Alexandru Vasluianu and Chao Ma and Danda Pani Paudel and Luc Van Gool and Radu Timofte},
  title     = {Single-model and Any-modality for Video Object Tracking},
  booktitle = {CVPR},
  pages     = {19156--19166},
  year      = {2024}
}

@inproceedings{vit,
  author    = {Alexey Dosovitskiy and Lucas Beyer and Alexander Kolesnikov and Dirk Weissenborn and Xiaohua Zhai and Thomas Unterthiner and Mostafa Dehghani and Matthias Minderer and Georg Heigold and Sylvain Gelly and Jakob Uszkoreit and Neil Houlsby},
  title     = {An Image is Worth 16x16 Words: Transformers for Image Recognition at Scale},
  booktitle = {ICLR},
  year      = {2021}
}

@inproceedings{vipt,
  author    = {Jiawen Zhu and Simiao Lai and Xin Chen and Dong Wang and Huchuan Lu},
  title     = {Visual Prompt Multi-modal Tracking},
  booktitle = {CVPR},
  pages     = {9516--9526},
  year      = {2023}
}

@inproceedings{wuyang2023dropmae,
  author    = {Qiangqiang Wu and Tianyu Yang and Ziquan Liu and Baoyuan Wu and Ying Shan and Antoni B. Chan},
  title     = {{DropMAE}: Masked Autoencoders with Spatial-Attention Dropout for Tracking Tasks},
  booktitle = {CVPR},
  pages     = {14561--14571},
  year      = {2023}
}

@inproceedings{Zhang_CVPR22_VTUAV,
  author    = {Pengyu Zhang and Jie Zhao and Dong Wang and Huchuan Lu and Xiang Ruan},
  title     = {Visible-Thermal {UAV} Tracking: A Large-scale Benchmark and New Baseline},
  booktitle = {CVPR},
  pages     = {8886--8895},
  year      = {2022}
}

@inproceedings{zhu2023rgbd1k,
  author    = {Xue-Feng Zhu and Tianyang Xu and Zhangyong Tang and Zucheng Wu and Haodong Liu and Xiao Yang and Xiao-Jun Wu and Josef Kittler},
  title     = {{RGBD1K}: A Large-scale Dataset and Benchmark for {RGB-D} Object Tracking},
  booktitle = {AAAI},
  volume    = {37},
  pages     = {3870--3878},
  year      = {2023}
}

@article{zhu2025adaptive,
  author    = {Zhu, Zhiruo and Zhong, Bineng and Liang, Qihua and Yang, Hongtao and Zheng, Yaozong and Li, Ning},
  title     = {Adaptive Expert Decision for RGB-T Tracking},
  journal   = {IEEE TCSVT},
  volume    = {35},
  number    = {10},
  pages     = {10330-10338},
  year      = {2025}
}

@article{9364880,
  author    = {Pengyu Zhang and Jie Zhao and Chunjuan Bo and Dong Wang and Huchuan Lu and Xiaoyun Yang},
  title     = {Jointly Modeling Motion and Appearance Cues for Robust RGB-T Tracking},
  journal   = {IEEE TIP},
  volume    = {30},
  pages     = {3335--3347},
  year      = {2021}
}

@inproceedings{cai2025spmtrack,
  author    = {Wenrui Cai and Qingjie Liu and Yunhong Wang},
  title     = {SPMTrack: Spatio-Temporal Parameter-Efficient Fine-Tuning with Mixture of Experts for Scalable Visual Tracking},
  booktitle = {CVPR},
  pages     = {16871--16881},
  year      = {2025}
}

@inproceedings{zhu2019dense,
  author    = {Yabin Zhu and Chenglong Li and Bin Luo and Jin Tang and Xiao Wang},
  title     = {Dense Feature Aggregation and Pruning for RGBT Tracking},
  booktitle = {ACM MM},
  pages     = {465--472},
  year      = {2019}
}

@inproceedings{gao2019deep,
  author    = {Yuan Gao and Chenglong Li and Yabin Zhu and Jin Tang and Tao He and Futian Wang},
  title     = {Deep Adaptive Fusion Network for High Performance RGBT Tracking},
  booktitle = {ICCVW},
  pages     = {91--99},
  year      = {2019}
}

@inproceedings{long2019multi,
  author    = {Chenglong Li and Andong Lu and Aihua Zheng and Zhengzheng Tu and Jin Tang},
  title     = {Multi-adapter RGBT Tracking},
  booktitle = {ICCVW},
  pages     = {2262--2270},
  year      = {2019}
}

@inproceedings{zhang2019multi,
  author    = {Lichao Zhang and Martin Danelljan and Abel Gonzalez-Garcia and Joost Van De Weijer and Fahad Shahbaz Khan},
  title     = {Multi-modal Fusion for End-to-end RGB-T Tracking},
  booktitle = {ICCVW},
  pages     = {2252--2261},
  year      = {2019}
}

@inproceedings{wang2020cross,
  author    = {Chaoqun Wang and Chunyan Xu and Zhen Cui and Ling Zhou and Tong Zhang and Xiaoya Zhang and Jian Yang},
  title     = {Cross-modal Pattern-propagation for RGB-T Tracking},
  booktitle = {CVPR},
  pages     = {7064--7073},
  year      = {2020}
}

@inproceedings{li2020challenge,
  author    = {Chenglong Li and Lei Liu and Andong Lu and Qing Ji and Jin Tang},
  title     = {Challenge-aware RGBT Tracking},
  booktitle = {ECCV},
  pages     = {222--237},
  year      = {2020}
}

@inproceedings{xiao2022attribute,
  author    = {Yun Xiao and Mengmeng Yang and Chenglong Li and Lei Liu and Jin Tang},
  title     = {Attribute-based Progressive Fusion Network for RGBT Tracking},
  booktitle = {AAAI},
  volume    = {36},
  pages     = {2831--2838},
  year      = {2022}
}

@inproceedings{liu2023quality,
  author    = {Lei Liu and Chenglong Li and Yun Xiao and Jin Tang},
  title     = {Quality-aware RGBT Tracking via Supervised Reliability Learning and Weighted Residual Guidance},
  booktitle = {ACM MM},
  pages     = {3129--3137},
  year      = {2023}
}

@inproceedings{zhang2023efficient,
  author    = {Tianlu Zhang and Hongyuan Guo and Qiang Jiao and Qiang Zhang and Jungong Han},
  title     = {Efficient RGB-T Tracking via Cross-modality Distillation},
  booktitle = {CVPR},
  pages     = {5404--5413},
  year      = {2023}
}

@article{chen2024emoe,
  author    = {Yucheng Chen and Lin Wang},
  title     = {eMoE-Tracker: Environmental MoE-Based Transformer for Robust Event-Guided Object Tracking},
  journal   = {IEEE RAL},
  volume    = {10},
  number    = {2},
  pages     = {1393-1400},
  year      = {2025}
}

@inproceedings{xia2022vision,
  author    = {Zhuofan Xia and Xuran Pan and Shiji Song and Li Erran Li and Gao Huang},
  title     = {Vision Transformer with Deformable Attention},
  booktitle = {CVPR},
  pages     = {4794--4803},
  year      = {2022}
}

@inproceedings{tan2025you,
  author    = {Yuedong Tan and Jiawei Shao and Eduard Zamfir and Ruanjun Li and Zhaochong An and Chao Ma and Danda Pani Paudel and Luc Van Gool and Radu Timofte and Zongwei Wu},
  title     = {What You Have is What You Track: Adaptive and Robust Multimodal Tracking},
  booktitle = {ICCV},
  pages={3455--3465},
  year      = {2025}
}

@article{zhang2024amnet,
  author    = {Tianlu Zhang and Xiaoyi He and Qiang Jiao and Qiang Zhang and Jungong Han},
  title     = {AMNet: Learning to Align Multi-modality for RGB-T Tracking},
  journal   = {IEEE TCSVT},
  volume    = {34},
  number    = {8},
  pages     = {7386--7400},
  year      = {2024}
}

@inproceedings{xia2023unified,
  author    = {Jianqiang Xia and Dianxi Shi and Ke Song and Linna Song and Xiaolei Wang and Songchang Jin and Chenran Zhao and Yu Cheng and Lei Jin and Zheng Zhu and Jianan Li and Gang Wang and Junliang Xing and Jian Zhao},
  title     = {Unified Single-Stage Transformer Network for Efficient {RGB-T} Tracking},
  booktitle = {IJCAI},
  pages     = {1471--1479},
  year      = {2024}
}

@article{fan2024querytrack,
  author    = {Huijie Fan and Zhencheng Yu and Qiang Wang and Baojie Fan and Yandong Tang},
  title     = {QueryTrack: Joint-modality Query Fusion Network for RGBT Tracking},
  journal   = {IEEE TIP},
  volume    = {33},
  pages     = {3187--3199},
  year      = {2024}
}

@inproceedings{xiao2025cross,
  author    = {Yun Xiao and Jiacong Zhao and Andong Lu and Chenglong Li and Bing Yin and Yin Lin and Cong Liu},
  title     = {Cross-modulated Attention Transformer for {RGBT} Tracking},
  booktitle = {AAAI},
  volume    = {39},
  pages     = {8682--8690},
  year      = {2025}
}

@inproceedings{yang2022prompting,
  author    = {Jinyu Yang and Zhe Li and Feng Zheng and Ales Leonardis and Jingkuan Song},
  title     = {Prompting for Multi-modal Tracking},
  booktitle = {ACM MM},
  pages     = {3492--3500},
  year      = {2022}
}

@article{hendrycks2016gaussian,
  title={Gaussian Error Linear Units (GELUs)},
  author={Hendrycks, Dan and Gimpel, Kevin},
  journal={arXiv preprint arXiv:1606.08415},
  year={2016}
}

@article{elfwing2018sigmoid,
  title={Sigmoid-weighted linear units for neural network function approximation in reinforcement learning},
  author={Elfwing, Stefan and Uchibe, Eiji and Doya, Kenji},
  journal={NN},
  volume={107},
  pages={3--11},
  year={2018}
}

@inproceedings{liu2025mambavlt,
  author    = {Xinqi Liu and Li Zhou and Zikun Zhou and Jianqiu Chen and Zhenyu He},
  title     = {Mambavlt: Time-evolving multimodal state space model for vision-language tracking},
  booktitle = {CVPR},
  pages     = {8731--8741},
  year      = {2025}
}

@inproceedings{guo2025dreamtrack,
  title={DreamTrack: Dreaming the Future for Multimodal Visual Object Tracking},
  author={Mingzhe Guo and Weiping Tan and Wenyu Ran and Liping Jing and Zhipeng Zhang},
  booktitle={CVPR},
  pages={7201--7210},
  year={2025}
}

@inproceedings{feng2025atctrack,
  author    = {Xiaokun Feng and Shiyu Hu and Xuchen Li and Dailing Zhang and Meiqi Wu and Jing Zhang and Xiaotang Chen and Kaiqi Huang},
  title     = {ATCTrack: Aligning Target-Context Cues with Dynamic Target States for Robust Vision-Language Tracking},
  booktitle = {ICCV},
  pages={19850--19861},
  year      = {2025}
}

@article{zhu2024unimod1k,
  title={UniMod1K: Towards a more universal large-scale dataset and benchmark for multi-modal learning},
  author={Zhu, Xue-Feng and Xu, Tianyang and Liu, Zongtao and Tang, Zhangyong and Wu, Xiao-Jun and Kittler, Josef},
  journal={IJCV},
  volume={132},
  number={8},
  pages={2845--2860},
  year={2024},
  publisher={Springer}
}

@article{xu2024learning,
  title={Learning adaptive spatio-temporal inference transformer for coarse-to-fine animal visual tracking: algorithm and benchmark},
  author={Xu, Tianyang and Kang, Ze and Zhu, Xuefeng and Wu, Xiao-Jun},
  journal={IJCV},
  volume={132},
  number={7},
  pages={2698--2712},
  year={2024},
  publisher={Springer}
}

@article{zhu2024self,
  title={Self-supervised learning for RGB-D object tracking},
  author={Zhu, Xue-Feng and Xu, Tianyang and Atito, Sara and Awais, Muhammad and Wu, Xiao-Jun and Feng, Zhenhua and Kittler, Josef},
  journal={PR},
  volume={155},
  pages={110543},
  year={2024},
  publisher={Elsevier}
}

@article{hu2025adaptive,
  title={Adaptive perception for unified visual multi-modal object tracking},
  author={Hu, Xiantao and Zhong, Bineng and Liang, Qihua and Shi, Liangtao and Mo, Zhiyi and Tai, Ying and Yang, Jian},
  journal={IEEE TAI},
  year={2025},
  volume={6},
  number={10},
  pages={2819-2829},
  publisher={IEEE}
}

\end{document}